\pdfoutput=1

\documentclass[11pt]{article}

\usepackage[]{EMNLP2023}

\usepackage{times}
\usepackage{latexsym}

\usepackage[T1]{fontenc}

\usepackage[utf8]{inputenc}

\usepackage{microtype}

\usepackage{inconsolata}
\usepackage{booktabs}
\usepackage{enumitem}
\usepackage{fdsymbol}
\usepackage{graphicx}
\usepackage{mathtools}
\usepackage{microtype}
\usepackage{multirow}
\usepackage{subfigure}
\usepackage{xspace}

\usepackage{svg}
\usepackage{subfigure}

\newcommand{\our}{\mbox{\textsc{DocMaster}}\xspace}

\newcommand\blfootnote[1]{%
  \begingroup
  \renewcommand\thefootnote{}\footnote{#1}%
  \addtocounter{footnote}{-1}%
  \endgroup
}

\title{\our: A Unified Platform for Annotation, \\Training, \& Inference in Document Question-Answering}

\author{
  Alex Nguyen$^{\diamondsuit}$ $\quad$ Zilong Wang$^{\diamondsuit}$ $\quad$ Jingbo Shang$^{\diamondsuit, \heartsuit, \spadesuit}$ $\quad$  Dheeraj Mekala$^{\diamondsuit, \spadesuit}$ \\
  $^\diamondsuit$University of California San Diego\\
  $^\heartsuit$ Hal\i c\i o\u glu Data Science Institute, University of California San Diego\\
  \small \texttt{\{atn021, zlwang, jshang, dmekala\}@ucsd.edu}
}

\begin{document}
\maketitle

\begin{abstract}
    The application of natural language processing models to PDF documents is pivotal for various business applications yet the challenge of training models for this purpose persists in businesses due to specific hurdles. These include the complexity of working with PDF formats that necessitate parsing text and layout information for curating training data and the lack of privacy-preserving annotation tools.
    This paper introduces \our, a unified platform designed for annotating PDF documents, model training, and inference, tailored to document question-answering. 
    The annotation interface enables users to input questions and highlight text spans within the PDF file as answers, saving layout information and text spans accordingly. 
    Furthermore, \our supports both state-of-the-art layout-aware and text models for comprehensive training purposes.
    Importantly, as annotations, training, and inference occur on-device, it also safeguards privacy.
    The platform has been instrumental in driving several research prototypes concerning document analysis such as the AI assistant utilized by University of California San Diego's (UCSD) International Services and Engagement Office (ISEO) for processing a substantial volume of PDF documents. 
    \blfootnote{$\spadesuit$ Corresponding Authors}
\end{abstract}
\begin{figure}[t]
    \centering
    \includegraphics[width=0.48\textwidth]{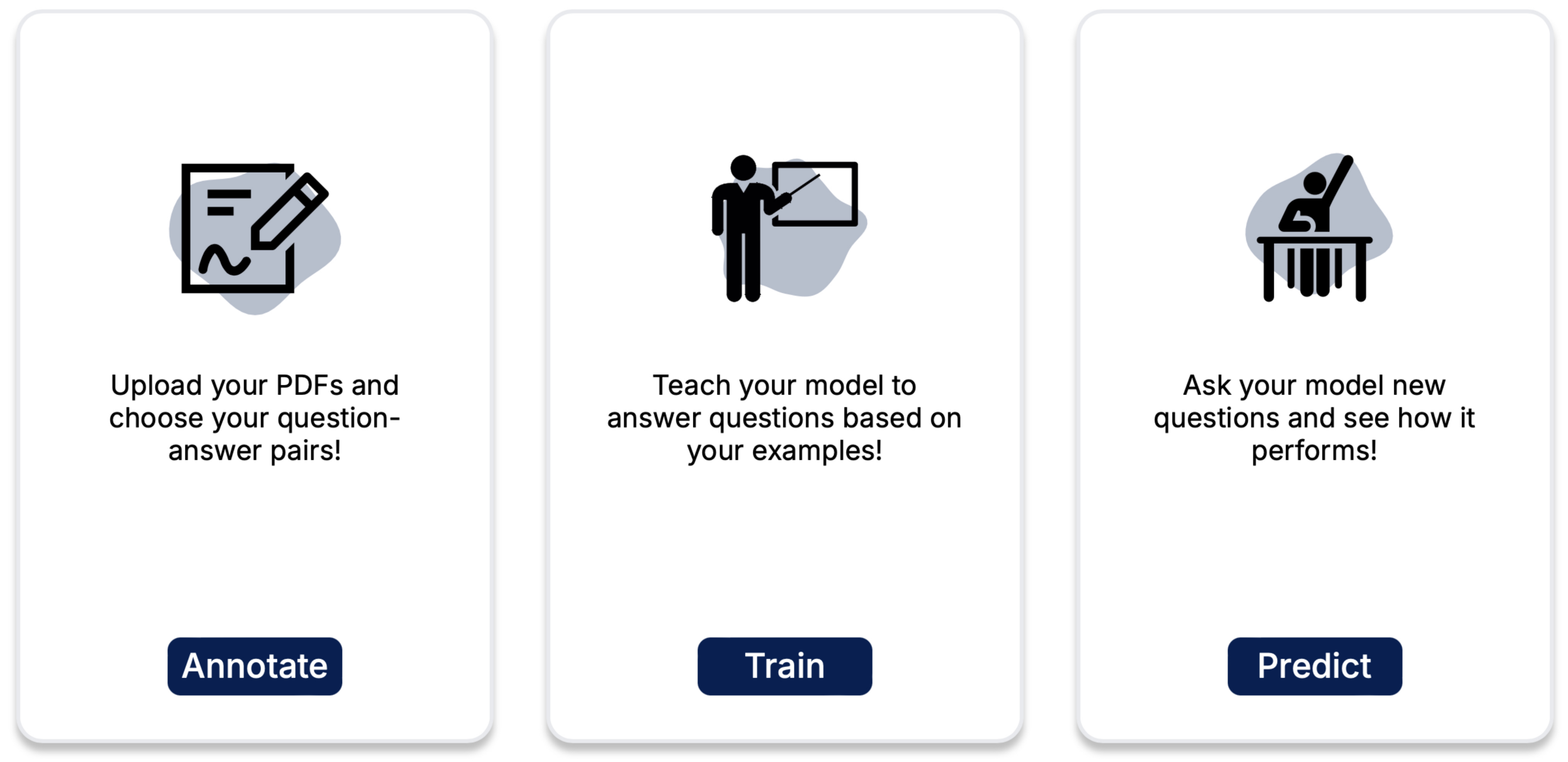}
    \caption{\our supports annotation, model training, and inference functionalities for document question-answering in a single platform.}
    \label{fig:overview}
\end{figure}

\section{Introduction}
Documents and forms are omnipresent within enterprises encompassing financial bills like invoices, purchase records, financial statements, and official communications such as notices and announcements.
The application of machine learning for automating document processing stands to significantly accelerate processing times~\cite{tan2023information}.

Visually rich document understanding has recently attracted much attention from researchers.
A simple approach involves parsing text from PDFs and leveraging established Natural Language Processing (NLP) models~\cite{Devlin2019BERTPO, liu2019roberta, NEURIPS2020_1457c0d6}. However, these methods overlook the valuable layout information embedded within PDFs.
A series of works have been done to incorporate the layout features into the pre-training framework. LayoutLM~\citep{xu2020layoutlm} first proposes to encode the spatial relationships of words by embedding their position coordinates in an embedding layer. Following this direction, \citet{xu2021layoutlmv2,huang2022layoutlmv3} move beyond basic embedding techniques and specifically adapt the attention layers of the Transformer architecture to model the relative positional relationships within the 2D space of document pages.
\citet{gu2021unidoc,wang2022mgdoc}, on the other hand, purse a more comprehensive understanding of the layout structure. They achieve this by encoding the hierarchical relation in the documents.


Despite the availability of these models, the persistent challenge lies in training them with custom business data due to particular obstacles. Firstly, working with the intricacies of the PDF format proves to be a nontrivial task~\cite{lo2023papermage}. PDFs store text as character glyphs along with their positions on a page, necessitating complex operations to convert this data into usable text for NLP models.
Operations like inferring token boundaries and managing white spacing are error-prone and add to the complexity. 
Secondly, organizations frequently handle sensitive documents that demand in-house tools for annotating and curating training data.


Addressing these obstacles, we introduce \our, a unified platform designed for annotating PDF documents, model training, and inference for the question-answering (QA) task, as shown in Figure~\ref{fig:overview}. 
The annotation interface is designed to maintain the layout integrity, requiring users to upload PDFs, provide questions, and highlight their specific text spans as answers in the PDFs. 
Once identified, it processes the PDF content, saving both textual and layout details. 
Privacy measures involve on-device processing, eliminating reliance on third-party services, and securely storing annotations within a local database.
\our accommodates an extensive array of models, encompassing both layout-aware models like LayoutLM~\cite{xu2020layoutlm} and text-only ones such as RoBERTa~\cite{liu2019roberta}. 
The inference interface is user-friendly and accepts a PDF document and a trained model. 
It simplifies the task of locating answers to specific questions by highlighting relevant spans within the PDF document.



We deployed \our in a practical scenario within the ISEO at UCSD, addressing the processing of hundreds of supporting documents for students to issue their work permits. 
Previously, staff members engaged in the manual review of each document before approving work permits. 
Through \our, they annotated and trained a QA model seamlessly, leading to a remarkable seven-fold increase in the average number of documents processed per hour.

We present video demonstration, live demo website, and code on the project webpage.\footnote{\url{https://alextongdo.github.io/doc-master-webpage/}}.
\section{Related Work}

\paragraph{Language Modeling with PDFs}

PDFs are widely used in daily life. It is trivial to resort to traditional language models, such as BERT~\citep{Devlin2019BERTPO}, RoBERTa~\citep{liu2019roberta}, and T5~\citep{raffel2020exploring}, to automatically understand the document contents. However, unlike the pure-text documents~\cite{mekala-etal-2022-lops}, PDFs carry rich information not only through the textual contents but also via the rich layout structure, presenting challenges for language models to comprehensively understand their contents. \citet{xu2020layoutlm,hong2022bros,garncarek2021lambert} propose to use the coordinates of words in the page as the representation for the layout structure. They embed the coordinates in the embedding layer and add relative weights in the self-attention layers. \citet{xu2021layoutlmv2,huang2022layoutlmv3} incorporate the visual features from the document images. Following the previous works, \citet{tang2023unifying,lv2023kosmos,perot2023lmdx} enlarge the scale of pre-training and improve language models capability in understanding PDFs of various formats.

\paragraph{Systems for Document AI}

Document AI is drawing significant interest from both academia and industry. In addition to various language modeling techniques, major companies have also launched their proprietary Document AI services, including Google Cloud~\footnote{\url{https://cloud.google.com/document-ai}}, Microsoft Azure~\footnote{\url{https://azure.microsoft.com/en-us/products/ai-services/ai-document-intelligence}}, Amazon Web Services~\footnote{\url{https://aws.amazon.com/machine-learning/ml-use-cases/document-processing/fintech/}}, etc. 
Although proprietary systems offer convenient and stable services, they are primarily business-oriented and lack transparency for those outside the company. Additionally, there are non-commercial Document AI systems available, such as \citet{lo2023papermage,bryan2023efficientocr}. However, none of these systems comprehensively enable users to combine annotation, training, and inference within a single system. 
In contrast, \our allows users to navigate the entire pipeline of Document-QA task, successfully eliminating programming barriers that hinder general users from utilizing Document AI tools.

\section{\our: Design}

\begin{figure*}[t]
  \centering
  \includegraphics[width=0.85\textwidth]{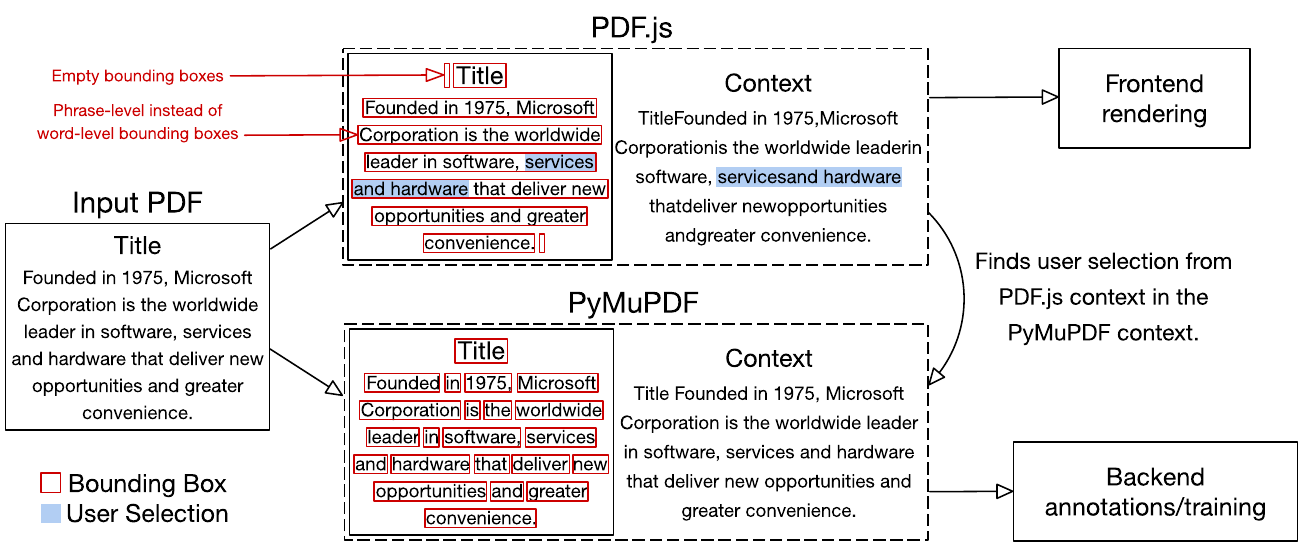}
  \caption{Training and inference with layout-aware models requires a bounding box for each word. 
  PDF.js cannot reliably provide this data because of its phrase-level bounding boxes instead of word-level and empty bounding boxes.
  PyMuPDF solves this issue, but the text parsed by PDF.js and PyMuPDF can differ. \our uses PDF.js for frontend rendering and PyMuPDF in the backend and provides a robust method for mapping a PDF.js selection to the PyMuPDF context.}
  \label{fig:annotation-challenges}
\end{figure*}

This section delves into the design aspects of our platform. 
\our has three interfaces: (1) The Annotation interface, which processes a zip file containing PDF documents, enabling user annotation through text highlighting. 
(2) The Training interface, facilitating the training of both layout-aware and text models. 
(3) The Inference interface, which accepts a set of documents as input, allows users to select their trained model and highlights predictions on the PDFs.
The \our application is intended to run on the organization's servers. As such, it is configured to automatically set up and run multiple Docker containers, enabling portability across environments.

\begin{figure}[t]
    \centering
    \includegraphics[width=0.48\textwidth]{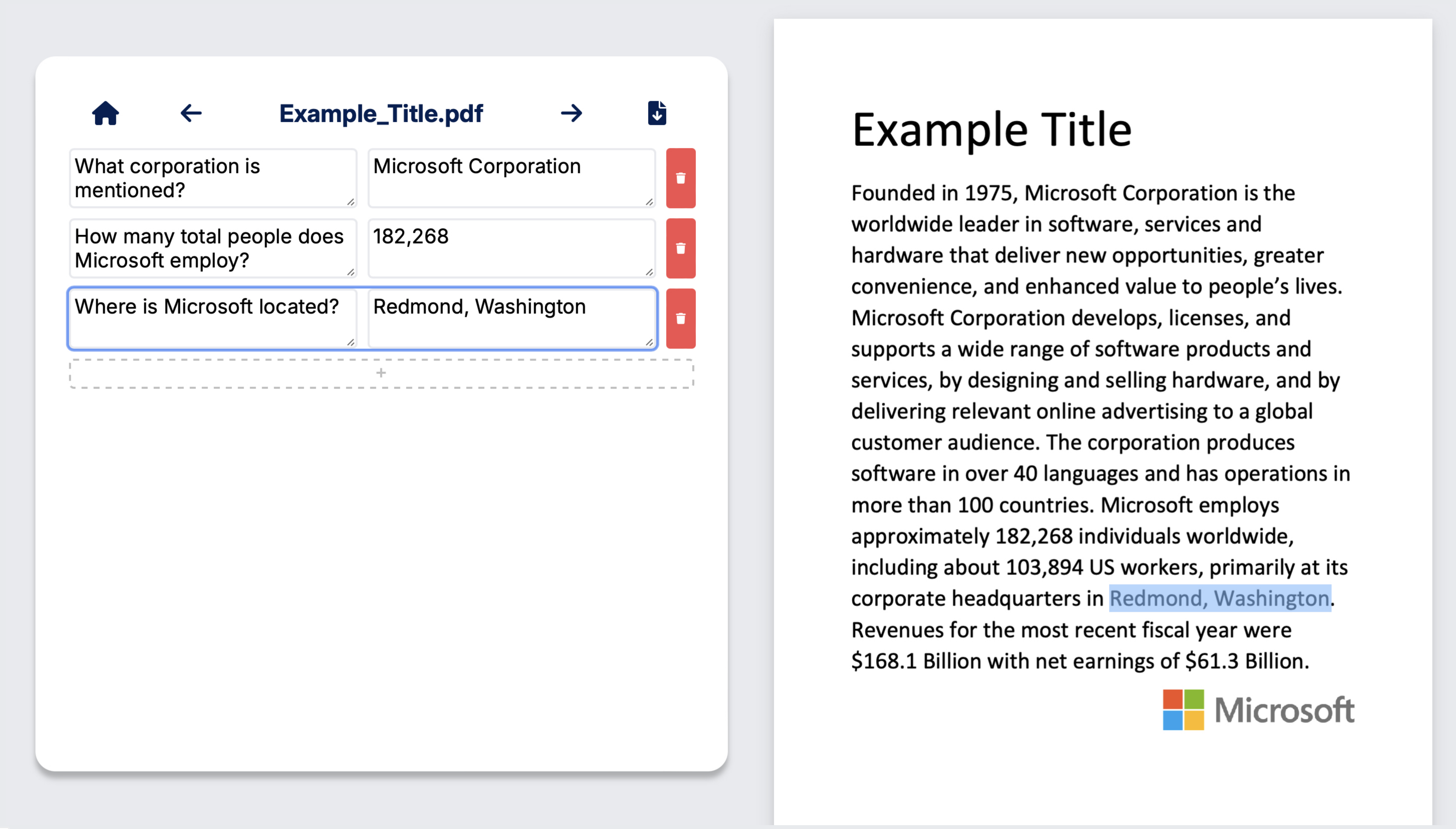}
    \caption{The annotation interface of \our. The users upload a PDF/a zip of PDFs, input their questions and highlight the answers in each PDF.}
    \label{fig:annotation}
\end{figure}

\subsection{Annotation}



The annotation interface streamlines the user process of uploading PDFs, and inputting questions and their corresponding answers, achieved through highlighting relevant text spans within the PDF.
For layout models, it is essential to capture the layout information of the highlighted span.
Consequently, the annotation interface must fulfill three essential requirements: (1.) accurately display the PDF, (2.) enable text highlighting, and (3.) collect layout information of the highlighted span.


We utilize Mozilla PDF.js\footnote{\url{https://mozilla.github.io/pdf.js/}} to embed the input document as a canvas onto the webpage, providing an engaging frontend experience. 
PDF.js incorporates an invisible \textit{textlayer}, enabling selectable text on the canvas, enhancing the user interface. 
Despite its advantages, the textlayer's bounding box information, which offers layout details, presents several challenges. 
Firstly, it provides bounding box information primarily for spans determined by PDF.js, often encompassing entire lines and phrases but not consistently individual words. 
For example, in Figure~\ref{fig:annotation-challenges}, the user selects "\textit{services and hardware}" and PDF.js provides bounding box information for "\textit{leader in software, services}" and "\textit{and hardware that deliver new}" separately, making it challenging to obtain the bounding box information for the selected text.
Secondly, it occasionally detects empty spans and provides irrelevant bounding box information as shown in Figure~\ref{fig:annotation-challenges}. 
Finally, the accuracy of highlighted text detected through PDF.js is not always reliable and is susceptible to whitespace errors, as illustrated by the user selection of "\textit{services and hardware}" in Figure~\ref{fig:annotation-challenges}, where spaces in the middle were not accurately preserved.

To address these limitations, we employ PyMuPDF\footnote{\url{https://pymupdf.readthedocs.io/en/latest/}} on the backend, a Python library that consistently provides word-level bounding boxes with accuracy. 
While PyMuPDF excels in providing accurate bounding box information, it cannot render PDFs on the webpage, hindering user-friendly text highlighting. 
Consequently, we integrate PDF.js in the frontend and PyMuPDF in the backend, leveraging the strengths of both. 
However, this integration introduces a compatibility challenge, requiring the conversion of user selections from PDF.js context to the corresponding selections in the PyMuPDF context.

If the user-selected text is uniquely identifiable within the PDF.js context, locating its position in the PyMuPDF context is straightforward. However, when dealing with non-unique selections, we encounter the challenge of distinguishing among multiple substrings in the PyMuPDF context that could potentially represent the desired selection. To address this, we leverage the bounding box information provided by PDF.js, which often corresponds to the sentence or phrase containing the selected text. This allows us to narrow down the search area and focus specifically on that text for accurate identification.
An example annotation interface is shown in Figure~\ref{fig:annotation}.


\begin{figure}[t]
    \centering
    \includegraphics[width=0.48\textwidth]{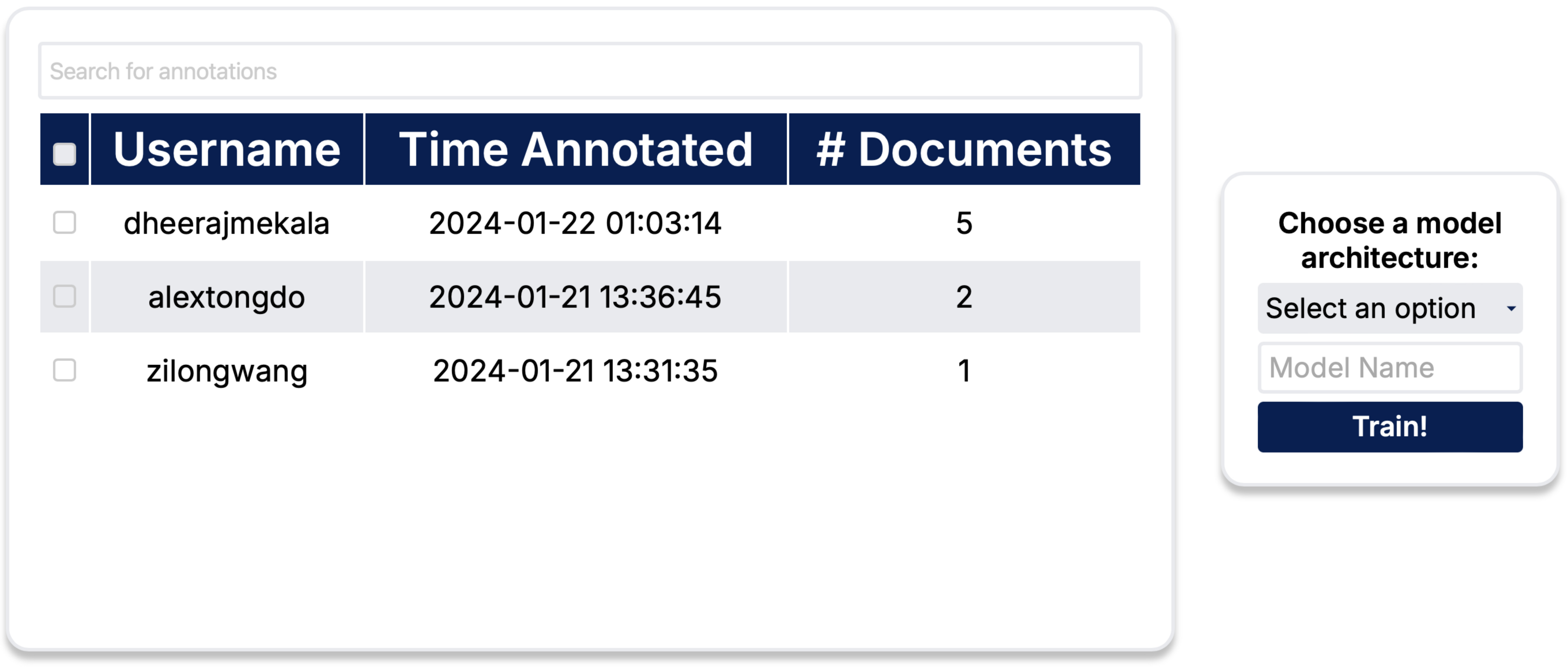}
    \caption{In training interface, the users can select one of the base models and train it using the previously annotated documents. Each row in the table indicates an annotation session and shows the number of documents annotated during that session.}
    \label{fig:training}
\end{figure}

\subsection{Training}

After annotating their desired PDFs, users can pick and choose which annotations they would like to include as training data. 
To manage the potential high influx of PDFs, \our organizes documents into sessions, affording users the choice to either fully include or exclude entire sessions. 
A new session is generated each time a user logs in, consolidating all annotations made during the active browsing window. 
Should modifications be necessary for an already annotated document, re-uploading a previously annotated PDF retrieves and removes its data from the prior session, enabling the updated data to be stored in a new session.

The training interface is shown in Figure~\ref{fig:training}.
In the training interface, all sessions are presented, showcasing the number of annotated documents and the corresponding times of annotation. 
This display streamlines the data selection process, providing transparency and accessibility. 
This information is shared publicly on the locally hosted \our platform, fostering collaborative efforts within a team. 
Consequently, any user can leverage annotations performed by others to train a model, promoting team-wide collaboration.

\our uses the transformers library from Huggingface~\cite{wolf2019huggingface} for in-house training and inference. Annotations and trained model weights are saved in a local SQL database, eliminating dependence on third-party services and preserving data privacy.

\subsection{Inference}

The inference interface enables users to choose their preferred trained model and submit a set of documents for predictions. 
As \our already leverages layout information in the annotation interface, we extend this approach to enhance user experience in the inference interface. 
Specifically, when users upload a new set of PDFs and questions to their QA model, \our not only provides the inferred text but also a copy of the input PDF with highlighted bounding boxes corresponding to the inference. 
This highlighting aids users in pinpointing the location of their answers and any relevant surrounding context. 
Additionally, users can conveniently download the highlighted PDFs for future reference.

\section{\our: Building YOUR Document QA System}

How can an organization utilize \our to implement a document QA system tailored to their use case? In this section, we illustrate a hypothetical scenario where the HR department of a company seeks to improve its onboarding process through the integration of a QA system.


\paragraph{Privacy-preserving}

Shang Data Lab, Inc. has an HR team aiming to implement a QA system for new hires to address queries related to various onboarding documents. 
However, due to the sensitive nature of these documents, the HR team is cautious about utilizing third-party services considering potential data leakage~\cite{nasr2023scalable}. 
Their preference is to ensure internal documents never leave their servers. 
Recognizing the open-sourced system \our for its emphasis on privacy, Shang Data Lab, Inc. finds it to be a suitable solution meeting their specific requirements.

\paragraph{Ease of Deployment}

Setting up \our is straightforward, involving the cloning of source code and the execution of a single command: ``\texttt{docker compose up}''. 
Leveraging Docker, a widely adopted containerization software, Shang Data Lab, Inc. can swiftly have their own \our operational within a few minutes.

\paragraph{Parallel Annotation}

Intending to train a QA model to aid in comprehending onboarding documents, the HR team at Shang Data Lab, Inc. allocates tasks to each team member, requiring them to generate questions for a subset of documents to curate training data. Utilizing \our, each team member logs in and uploads a few onboarding PDFs to the annotation interface. 
Within this interface, they can annotate the answers to their questions by highlighting relevant text in the PDF. 
Working concurrently, the HR team successfully compiles a training dataset containing multiple questions and corresponding answers relevant to each onboarding document.

\paragraph{Training \& Inference}


With the newly curated dataset, Shang Data Lab, Inc. initiates the training of QA models seamlessly through the training interface. 
Utilizing the platform's features, they have the flexibility to opt for training either a text-only or layout-aware model. Once the model is trained, they can deploy it using the inference interface, enabling new hires to leverage its capabilities.
New hires can easily upload a PDF and input a set of questions, receiving not only accurate answers but also benefiting from the ability to precisely locate the answers within the document through highlighted references. 

This scenario highlights the versatility of \our and its aptitude to address specific needs within the AI-as-a-service ecosystem.
\begin{figure}[t]
    \center
    \includegraphics[width=\linewidth]{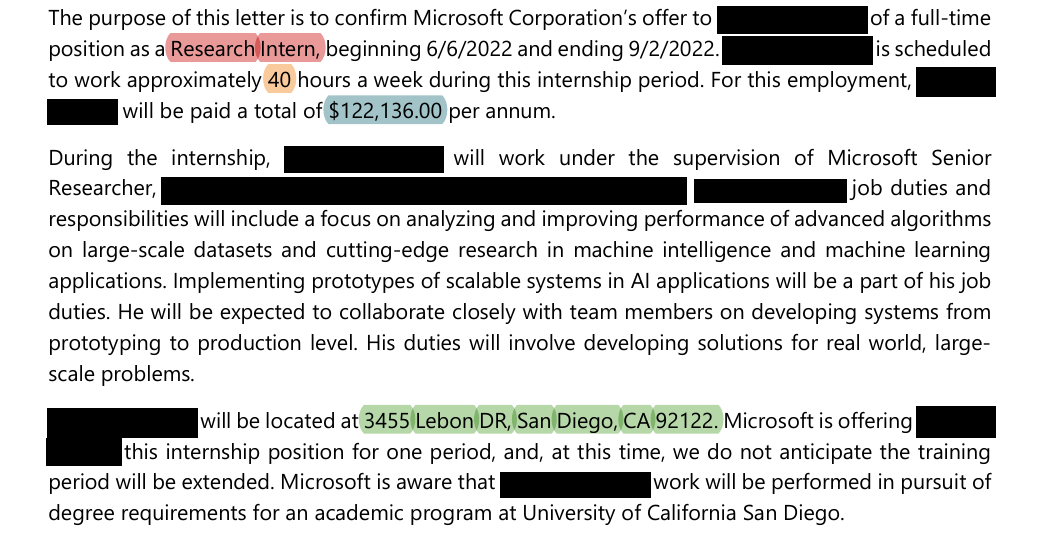}
    \caption{
    Highlighted answers for questions asked by ISEO office staff on a supporting document. The questions are: ``\textit{What is the job title?}'' \textcolor[HTML]{cf5a5a}{(red)}, ``\textit{What are the work hours per week?}'' \textcolor[HTML]{f1a05f}{(orange)}, ``\textit{What is the salary or hourly rate?}'' \textcolor[HTML]{66979f}{(blue)}, ``\textit{Where is the internship address?}'' \textcolor[HTML]{7fb56d}{(green)}. Private information is redacted.} 
    \label{fig:example_iseo}
\end{figure}

\section{Public Deployment: Takeaways \& Testimonials}

The ISEO at UCSD\footnote{\url{https://ispo.ucsd.edu/}} oversees immigration services for international students. This responsibility encompasses tasks such as certifying students' admission to full-time study programs, issuing work permits, and managing various other related processes. Each certification request undergoes a meticulous manual review of its accompanying supporting documents. During peak periods, the volume of applications can reach into the thousands.

Presently, each request is processed manually, involving a staff member who reviews supporting documents, communicates with relevant sub-teams for additional assessment, and ultimately electronically approves or declines the request. 
The manual nature of this process is labor-intensive and demands significant human effort. 
Furthermore, any delay in processing requests poses potential challenges for international students, including leaving the country or delays in commencing employment.


To tackle this issue, we deployed \our to streamline the review process, focusing on a specific scenario: the issuance of work permits for internships, as a prototype use case. 
Traditionally, ISEO staff manually verifies essential fields and grants approval for work permits upon the submission of supporting documents by students. 
The eight key fields subject to review include employer details, salary information, job description, supervisor name \& email address, weekly work hours, internship location, and start \& end dates. 


We formulate this as a QA task and train a model to extract necessary fields~\cite{mekala-etal-2022-leveraging}. 
To generate training data, four individuals without a machine learning background utilize \our's annotation interface to annotate five documents each. 
Each annotator formulates a question for every required field and highlights the relevant answer span within the PDF~\cite{mekala-etal-2023-zerotop}.
The collected annotations, encompassing both text and layout information, are aggregated. 
Subsequently, we train two models, RoBERTa-base and LayoutLM-base, for three epochs using this annotated dataset. 
During the inference phase, new student documents are uploaded to the interface, and the staff member inputs questions corresponding to the required fields. 
The user then selects the trained model, and answers for each field are highlighted within the PDF, as illustrated in Figure~\ref{fig:example_iseo}.


Our test set comprises 128 applications, encompassing a total of 1024 questions.
After consulting with the ISEO staff, we learned that traditional QA task metrics such as exact match accuracy (Acc), f1-score (F1) alone are not sufficient; the most crucial metric for them is the average processing time of a document.
The more easily identifiable the fields are, the quicker the document processing time becomes.
Consequently, we tailored our automated metrics to account for this priority.

We define our correctness (Corr) metric as follows to consider partial overlaps with the ground truth.
More precisely, we calculate the length of the longest contiguous matching subsequence and define a prediction as correct when the overlapping subsequence length exceeds 20\% of the prediction's total length. 
In cases where there is no overlap, we utilize Python's difflib SequenceMatcher\footnote{\url{https://docs.python.org/3/library/difflib.html}} to compute the longest contiguous matching subsequence between 
\(P\) and \(T\), excluding any ``junk''. 
A prediction is considered correct if the computed score is greater than 0.5; otherwise, it is deemed incorrect.
Mathematically,
\begin{align*}
    \small
   \text{Corr}( P,T) =\begin{cases}
    1 & \text{if} \ \frac{\text{len}( P_{\text{indexes}} \cap T_{\text{indexes}})}{\text{max}\left(\text{len}( P) ,\ \text{len}( T)\right)}  >0.2\\
    1 & \text{else if SequenceMatcher} ( P,T)  >0.5\\
    0 & \text{else}
    \end{cases} 
\end{align*}


We additionally incorporate the Euclidean distance between the predicted bounding box and the ground truth bounding box (Dist) as a performance metric. Recognizing the challenge posed by raw distance interpretation, we opt for a relative distance measurement, specifically, the distance normalized by the diagonal length of the page.
A shorter distance indicates an easier identification of ground truth, leading to reduced processing time.

\begin{table}[t]
    \center
    \caption{Performance Results on 128 applications test set in \%. 
    }
    \label{tbl:perf}
    \vspace{-3mm}
    \resizebox{\linewidth}{!}{
    \setlength{\tabcolsep}{2mm}{
    \begin{tabular}{c c c c c}
        \toprule
            {\textbf{Model}} & {\textbf{Acc}} & {\textbf{F1}} & {\textbf{Corr}} & {\textbf{Dist}} \\
        \midrule
        RoBERTa-base & 76.23 & 83.77 & 93.56 & 1.13\\
        LayoutLM-base & 75.98 & 83.07 & 93.36 & 1.86\\
        \bottomrule
    \end{tabular}
    }}
\end{table}

The performance results for RoBERTa-base and LayoutLM-base are detailed in Table~\ref{tbl:perf}. 
Notably, both models exhibit a similar performance on the test set, achieving a correctness score of approximately 94\%. 
The disparity between exact match accuracy and correctness scores underscores the inadequacy of standard academic evaluation metrics, prompting the need for a reevaluation of metrics tailored to real-life deployment scenarios.
Furthermore, we compute average bounding box distance for incorrect predictions alone, revealing values of 19.57\% for RoBERTa-base and 24.39\% for LayoutLM-base. 
This implies that when predictions are inaccurate, they tend to be in close proximity, typically within 20\% of the page size, indicating correct localization despite incorrect answers.

We also measure throughput on the test set by deploying \our on an AMD EPYC 7453 28-Core Processor (56 CPUs, base frequency of 2.75 GHz, boost frequency of up to 3.45 GHz).
We prioritize the lightweight nature of the RoBERTa-base model over LayoutLM-base and consider it for practical deployment.
Leveraging \our, the ISEO experienced a sevenfold enhancement in the number of supporting documents that can be reviewed per hour, escalating from 15 to 100.

Considering the sensitivity of the information contained in supporting documents, encompassing details like home and work addresses, salary information, and supervisor details, \our stands out as a fitting solution, guaranteeing the privacy of confidential data with on-device computing. 
Offering both high performance and convenience, with the ability to annotate data, train models, and make predictions all within a unified platform, \our emerges as the optimal open-sourced platform for such use cases.

\section{Conclusion}


This work introduces \our, a unified Document-QA platform designed for annotation, training, and inference while prioritizing privacy preservation. 
\our aims to empower users to train and deploy their models for document QA purposes. 
Despite the availability of various models, there is a scarcity of open-sourced annotation platforms. 
Addressing this gap, \our is presented as an open-source solution where users can annotate PDFs effortlessly by simply highlighting relevant text. The platform has demonstrated its efficacy in constructing UCSD ISEO's AI assistant, contributing to a noteworthy seven-fold reduction in document processing time. 
The open-sourcing of \our is intended to empower businesses that necessitate in-house document QA platforms.

\section{Ethical Considerations}
We introduce a privacy-preserving document-QA platform and identify no ethical concerns associated with its use.

\section{Acknowledgments}
The authors thank Bryant Tan, Gilen Wu-hou, Jinya Jiang, Khai Luu for their valuable contributions. 
We also thank Pauline DeGuzman and Emily Stewart for their support.
Finally, we thank Vaidehi Gupta and Mai ElSherief for their guidance.

\clearpage\newpage

\bibliography{anthology,custom}

\newpage
\appendix

\clearpage\newpage

\end{document}